\documentclass[10pt,conference]{IEEEtran}
\IEEEoverridecommandlockouts

\usepackage[utf8]{inputenc}
\usepackage{amsmath,amssymb}
\usepackage{graphicx}
\usepackage{booktabs}
\usepackage{siunitx}
\usepackage{hyperref}
\usepackage{microtype}
\usepackage{xcolor}
\usepackage{enumitem}
\usepackage{tikz}
\usetikzlibrary{positioning}
\hypersetup{
  hidelinks,
  pdftitle={Assessing Neuromorphic Computing for Fingertip Force Decoding from Electromyography},
  pdfauthor={A. Shahrooei, L. Arthur, O. Patel, D. Kamper},
  pdfkeywords={neuromorphic computing, high-density surface EMG, spiking neural network, motor unit decomposition, fingertip force, neural interface, rehabilitation robotics}
}

\title{Assessing Neuromorphic Computing for Fingertip Force Decoding from Electromyography%
\thanks{\textbf{License note:} \;Copyright 2025 The Authors. This work is licensed under CC BY 4.0.}%
\thanks{\textbf{Contact:} \;ashahro@ncsu.edu \quad ORCID: \url{https://orcid.org/0009-0006-0337-523X}. \textbf{Presentation note:} This short report expands a poster presented at IEEE EMBS NER 2025 and at two NC State venues (College of Engineering Applied AI Symposium; ECE Graduate Research Symposium, tied for Best Poster).}
}

\author{
\IEEEauthorblockN{A. Shahrooei\IEEEauthorrefmark{1}, L. Arthur\IEEEauthorrefmark{2}, O. Patel\IEEEauthorrefmark{2}, and D. Kamper\IEEEauthorrefmark{2}}
\IEEEauthorblockA{\IEEEauthorrefmark{1}Department of Electrical and Computer Engineering, NC State University, USA}
\IEEEauthorblockA{\IEEEauthorrefmark{2}Lampe Joint Department of Biomedical Engineering, NC State University / University of North Carolina at Chapel Hill, USA}
}

\begin{document}
\maketitle

\begin{abstract}
High-density surface electromyography (HD-sEMG) provides a noninvasive neural interface for assistive and rehabilitation control, but mapping neural activity to user motor intent remains challenging. We assess a spiking neural network (SNN) as a neuromorphic architecture against a temporal convolutional network (TCN) for decoding the fingertip force from the motor unit (MU) firing derived from HD-sEMG. Data were collected from a single participant (10 trials) with two forearm electrode arrays; MU activity was obtained via FastICA-based decomposition, and models were trained on overlapping windows with causal convolutions throughout. On held-out trials, the TCN achieved 4.44\% MVF RMSE (Pearson $r = 0.974$) while the SNN achieved 8.25\% MVF ( $r = 0.922$). While TCN was more accurate, the SNN remains attractive for low-power, event-driven inference and may close the gap with architectural and hyperparameter refinements.
\end{abstract}


\noindent\textbf{Clinical Relevance---} HD-sEMG-based force decoding enables more intuitive assistive and rehabilitation control. Neuromorphic models aim to deliver this capability at low power for wearable, real-time use.

\section{Introduction}
\begin{figure}[t]
  \centering
  \includegraphics[width=\linewidth]{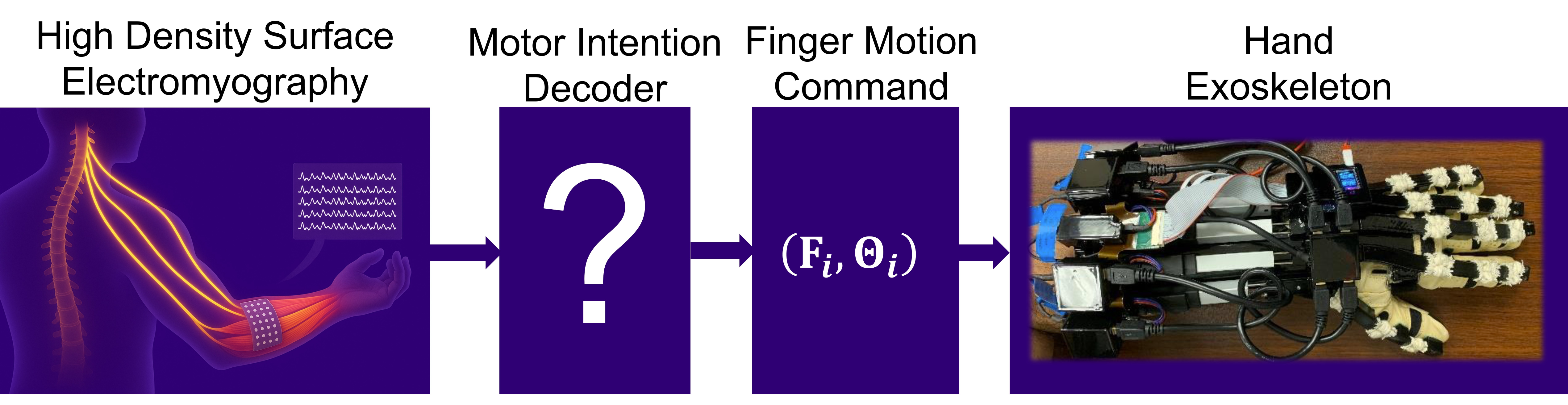}
  \caption{EMG-based neural interface for hand exoskeleton.}
  \label{fig:exoneuro}
\end{figure}

Loss of voluntary hand function affects a large population, particularly individuals post-stroke, limiting independence in activities of daily living. Assistive and rehabilitation devices (e.g., powered hand exoskeletons) offer strong potential to restore lost motor function, but their effectiveness depends on reliable, low-latency decoding of user intent. High-density surface electromyography (HD-sEMG) provides a noninvasive window into spinal motor neuron output with sufficient spatial resolution to separate motor unit (MU) activity. The missing piece is a robust mapping to continuously decode user intention (finger forces and joint angles) from HD-sEMG (Fig.~\ref{fig:exoneuro}).

Beyond raw EMG amplitudes, modern blind-source-separation techniques can decompose HD-sEMG into MU discharge timings (spike trains), effectively decoupling the neural drive to individual MUs. This noninvasive access to the \emph{spiking} output of spinal motor neurons enables interfaces that are closer to the underlying control signals delivered by the central nervous system and enables individual finger motion decoding.

A continuing challenge in fully exploiting the potential of HD-sEMG for neural interface development is reliable, accurate decoding \cite{Farina2025_ExtrationNeuralStrategies,ChenZhou2025_HDSEMG_TNSRE}. Prior work has explored convolutional and recurrent models (e.g., CNNs, LSTMs) for mapping sEMG to motion variables \cite{Fan2023_RobustDecodingCompBioMed,Yan2025_WristAngleTorque_TNSRE}. Spiking neural networks (SNNs) are attractive for low-power, event-driven computation and biologically inspired internal states; e.g., \cite{Tanzarella2023_TNSRE} decoded motor neuron activity from hand muscles. In this work, we examine an SNN versus a strong temporal convolutional network (TCN) baseline for predicting continuous fingertip force from HD-sEMG-derived MU activity.

In neuromorphic terms, SNNs process sparse spatio-temporal spike patterns directly and can be trained with surrogate gradients while maintaining time constants that reflect neuronal and synaptic dynamics; such models have achieved high accuracy on hand gesture decoding from MU spike trains \cite{Mansooreh_SNN} and are well suited for embedded, on-device inference with tight energy budgets. These attributes make SNNs compelling for wearable neural interfaces that must operate in real time with limited power (e.g., forearm bands or sleeve arrays). At the same time, for continuous regression of force, dilated-causal TCNs provide a strong baseline due to their ability to model long-range temporal dependencies with stable gradient flow. In this work, we therefore compare an SNN and a TCN within a common pipeline that starts from HD-sEMG, extracts MU activity, and predicts fingertip force from input features. In the remainder of the paper, we describe the recording and MU-extraction procedures, detail both models and the training setup, present quantitative results on held-out data, and discuss limitations and implications for neuromorphic control of hand exoskeletons.

\section{Methods}

\subsection{HD-sEMG Signal Generation \& Recording}
\begin{figure}[t]
  \centering
  \includegraphics[width=\linewidth]{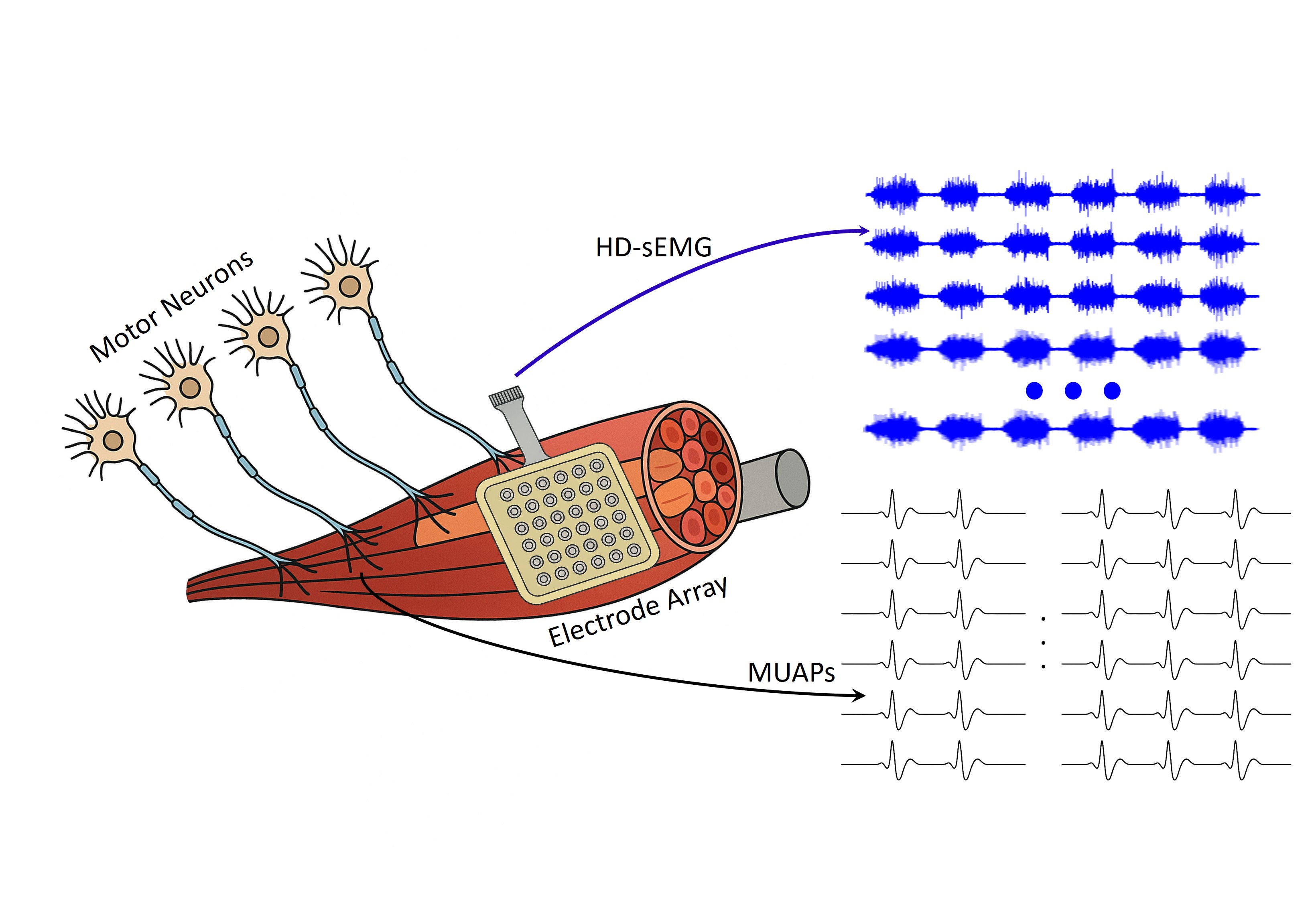}
  \caption{HD-sEMG signal generation and recording. Dense electrode arrays capture the superposition of MU action potentials over the forearm while fingertip force is measured with load cells.}
  \label{fig:hdsemg}
\end{figure}

Dense surface arrays record the spatiotemporal sum of MU action potentials (MUAPs) over the skin; higher electrode density increases the spatial resolution needed to separate individual MU activity \cite{ChenZhou2025_HDSEMG_TNSRE} (Fig.~\ref{fig:hdsemg}). In the present study, we used such arrays to sample activity from forearm flexor and extensor muscles while the participant generated controlled fingertip forces. For this work, data were collected from one neurotypical participant producing isometric index finger flexion to track a reference trapezoidal force trajectory with a plateau near 50\% MVF (maximum voluntary force). Two HD-sEMG arrays were used: 128-channel on the volar forearm and 64-channel on the dorsal forearm (OT Bioelettronica). Fingertip force was recorded using load cells. Trials lasted \(\sim\)30\,s; 10 trials were recorded for the task, following the same general protocol used in our multi-day reliability study \cite{Roy2025_EMBC_Reliability}.

During recording, the participant was seated comfortably facing a computer monitor that displayed the target force trajectory and their current fingertip force in real time. The right forearm was secured in a custom apparatus with adjustable brackets to maintain a neutral posture in pronation/supination and wrist flexion/extension. The index finger was positioned between a pair of six-axis load cells (Nano17 ATI Industrial Automation), such that flexion produced contact with the flexion load cell and extension produced contact with the opposing cell \cite{Roy2025_EMBC_Reliability}. Only the index finger flexion task was used for the present analysis. All experimental procedures were approved by the Institutional Review Board of North Carolina State University, and the participant provided written informed consent.

The 128-channel array was placed on the volar forearm centered over flexor digitorum superficialis, and the 64-channel array on the dorsal forearm centered over extensor digitorum communis, using the same anatomical landmarks and layout as in \cite{Roy2025_EMBC_Reliability}. Each array consisted of 3\,mm diameter electrodes arranged in an almost rectangular grid with 4\,mm inter-electrode spacing. HD-sEMG signals were amplified and bandpass filtered (10--900\,Hz) using an amplifier system (OT Bioelettronica) and sampled at 2048\,Hz; fingertip force signals from the load cells were sampled at the same rate using a National Instruments board. Acquisition was synchronized via a digital trigger.

\subsection{Extracting Motor Unit Activity}
Recorded EMG and force signals were synchronized using the trigger signal. EMG was filtered with a 60\,Hz notch and a 6th-order high-pass at 20\,Hz; force was low-pass filtered at 10\,Hz. After filtering, HD-sEMG channels were partitioned into subregions to account for spatial variation in signal quality. Channels were then selected by RMS/SNR within array subregions. 

Selected channels were whitened and decomposed using FastICA into MU source signals \cite{Hyvarinen2000_ICA,Negro2016_CBSS}. Formally, the HD-sEMG mixture $\mathbf{x}(t)\in\mathbb{R}^{C}$ is modeled as
\begin{equation}
    \mathbf{x}(t) = \mathbf{A}\,\mathbf{s}(t) + \boldsymbol{\varepsilon}(t),
\end{equation}
where $\mathbf{A}$ is an unknown mixing matrix, $\mathbf{s}(t)\in\mathbb{R}^{M}$ stacks MU source signals, and $\boldsymbol{\varepsilon}(t)$ is measurement noise. After centering and whitening,
\begin{equation}
    \mathbf{y}(t) = \mathbf{V}\bigl(\mathbf{x}(t)-\boldsymbol{\mu}\bigr),
\end{equation}
FastICA estimates an unmixing matrix $\mathbf{W}$ such that
\begin{equation}
    \hat{\mathbf{s}}(t) = \mathbf{W}\,\mathbf{y}(t)
\end{equation}
are statistically independent components interpreted as MU source signals.

For each source $i$, spike times $\{t_{i,k}\}$ were detected from $\hat{s}_i(t)$ and converted to a discrete binary spike train $s_i[n]\in\{0,1\}$, using peak detection followed by k-means++ clustering to separate spike and noise peaks. A continuous MU neural drive was then obtained by convolving $s_i[n]$ with a kernel $h[\ell]$,
\begin{equation}
    d_i[n] = \sum_{\ell=0}^{L-1} h[\ell]\, s_i[n-\ell],
\end{equation}
where $h[\ell]$ is a fixed smoothing kernel of length $L$. Population neural drives were formed as the sum of selected $d_i[n]$ across MUs within each muscle group (flexor/extensor). Duplicate and low-quality MUs were removed, retaining those with high clustering quality and strongest correlation to recorded force.

\begin{figure}[t]
  \centering
   \includegraphics[width=\linewidth]{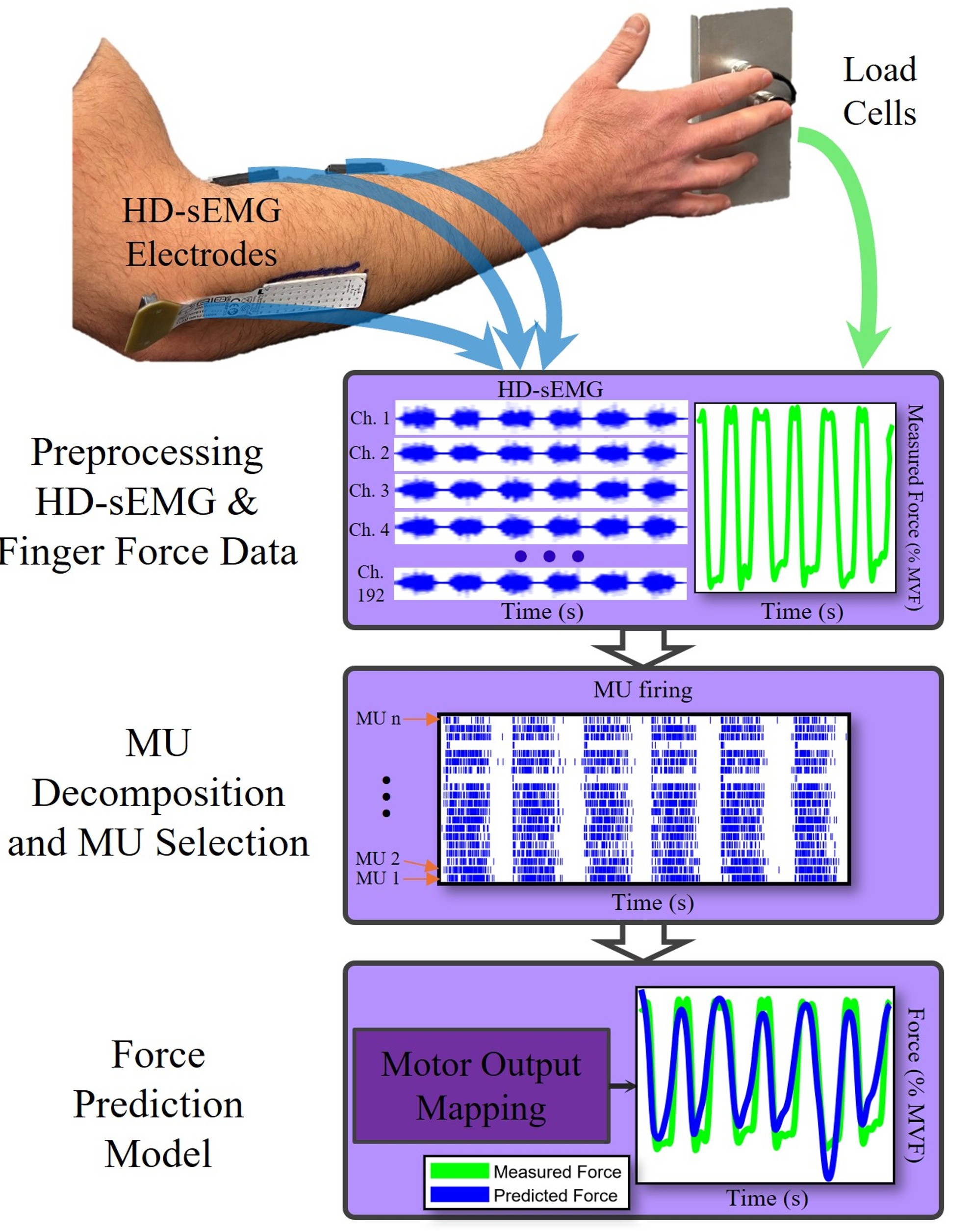}
  \caption{Neural interface development procedure: preprocessing, MU decomposition, feature formation, windowing, and causal target alignment.}
  \label{fig:procedure}
\end{figure}

Inputs were standardized (z-score using train-set statistics). Sliding windows of length \(T{=}256\) samples (200\,Hz) and stride 128 were used. To account for electromechanical delay, the force target was shifted by \(+80\) ms. Of the 10 trials, six were used for MU decomposition and training, two for validation, and two for testing (Fig.~\ref{fig:procedure}).

\subsection{Models and Training}
\textbf{Temporal Convolutional Network (TCN).} We adopt a causal TCN tailored for smooth continuous regression. A causal Conv1d stem (in\(\rightarrow\)64, kernel=9) is followed by six residual dilated blocks with dilations \([1,2,4,8,16,32]\). Each block applies Conv1d(64\(\rightarrow\)64, \(k{=}9\), dilation \(d_i\)) \(\rightarrow\) ReLU \(\rightarrow\) LayerNorm (over channel) \(\rightarrow\) Dropout \(p{=}0.1\) \(\rightarrow\) Conv1d(64\(\rightarrow\)64, \(k{=}9\), dilation \(d_i\)), with an identity skip connection. A causal \(1{\times}1\) Conv1d head maps 64\(\rightarrow\)1. All convolutions use causal-same padding to preserve time length.

Let \(\mathbf{x}\in\mathbb{R}^{T\times F}\) denote an input window (time \(T=256\), features \(F\)), and consider a generic 1-D convolution layer with kernel size \(k\), dilation \(d\), and weights \(\mathbf{W}\in\mathbb{R}^{C_{\text{out}}\times C_{\text{in}}\times k}\). For time index \(t\), the causal Conv1d output for channel \(c\) is
\begin{equation}
    y_c(t) = \sum_{j=1}^{C_{\text{in}}} \sum_{\ell=0}^{k-1}
    W_{c,j,\ell}\; x_j\bigl(t - d\,\ell\bigr),
\end{equation}
with padded values \(x_j(\tau)=0\) for \(\tau<0\) (causal-same padding keeps \(T\) unchanged). Within a residual block with dilation \(d_i\), the output \(\mathbf{z}^{(i)}(t)\) can be written as
\begin{equation}
    \mathbf{z}^{(i)}(t) = \mathbf{x}^{(i)}(t) + \mathcal{F}_{d_i}\bigl(\mathbf{x}^{(i)}\bigr)(t),
\end{equation}
where \(\mathcal{F}_{d_i}(\cdot)\) denotes the composition of Conv–ReLU–LayerNorm–Dropout–Conv with dilation \(d_i\). Stacking the stem and six residual blocks yields a large effective receptive field over the input window, while remaining strictly causal in time.

\textbf{Spiking Neural Network (SNN).} The SNN uses a causal Conv1d front-end (two layers; kernel 9; dilations 1 and 2; width 64) feeding a leaky-integrate-and-fire (LIF) spiking layer trained with a surrogate gradient. The membrane update uses a decay \(\beta_m{=}0.90\). Spikes are converted to rates with a causal synaptic low-pass readout \(y_t=\alpha y_{t-1} + (1-\alpha)s_t\) (learnable \(\alpha\)); a causal \(1{\times}1\) head maps 64\(\rightarrow\)1.

For each channel \(c\) in the LIF layer, the membrane potential \(v_t^{(c)}\) and spike output \(s_t^{(c)}\in\{0,1\}\) evolve as
\begin{align}
    v_t^{(c)} &= \beta_m\, v_{t-1}^{(c)} + I_t^{(c)} - v_{\text{th}}\, s_{t-1}^{(c)},\\
    s_t^{(c)} &= H\!\bigl(v_t^{(c)} - v_{\text{th}}\bigr),
\end{align}
where \(I_t^{(c)}\) is the input current from the conv front-end, \(v_{\text{th}}\) is a firing threshold, and \(H(\cdot)\) is the Heaviside step function. During training, the non-differentiable \(H(\cdot)\) is replaced by a surrogate derivative \(\partial s_t^{(c)}/\partial v_t^{(c)} \approx \phi(v_t^{(c)})\), with \(\phi(\cdot)\) a smooth function chosen for stable backpropagation. The synaptic low-pass readout implements an exponentially decaying memory of spikes,
\begin{equation}
    y_t^{(c)} = \alpha\, y_{t-1}^{(c)} + (1-\alpha)\, s_t^{(c)}, \qquad 0<\alpha<1,
\end{equation}
which is then mapped to force by the causal \(1{\times}1\) Conv1d head.

\textbf{Training \& evaluation.} Both models use MSE loss, Adam (lr \(=10^{-3}\)), batch size 32, and up to 80 epochs with validation early stopping. For a batch of \(N\) windows with targets \(y_{n,t}\) and predictions \(\hat{y}_{n,t}\), the loss is
\begin{equation}
    \mathcal{L}_{\text{MSE}} = \frac{1}{N T} \sum_{n=1}^{N} \sum_{t=1}^{T}
    \bigl(\hat{y}_{n,t} - y_{n,t}\bigr)^2.
\end{equation}
Metrics are RMSE (expressed as \%MVF) and Pearson correlation \(r\), averaged across held-out test trials. 

\begin{figure}[t]
  \centering
   \includegraphics[width=0.98\linewidth]{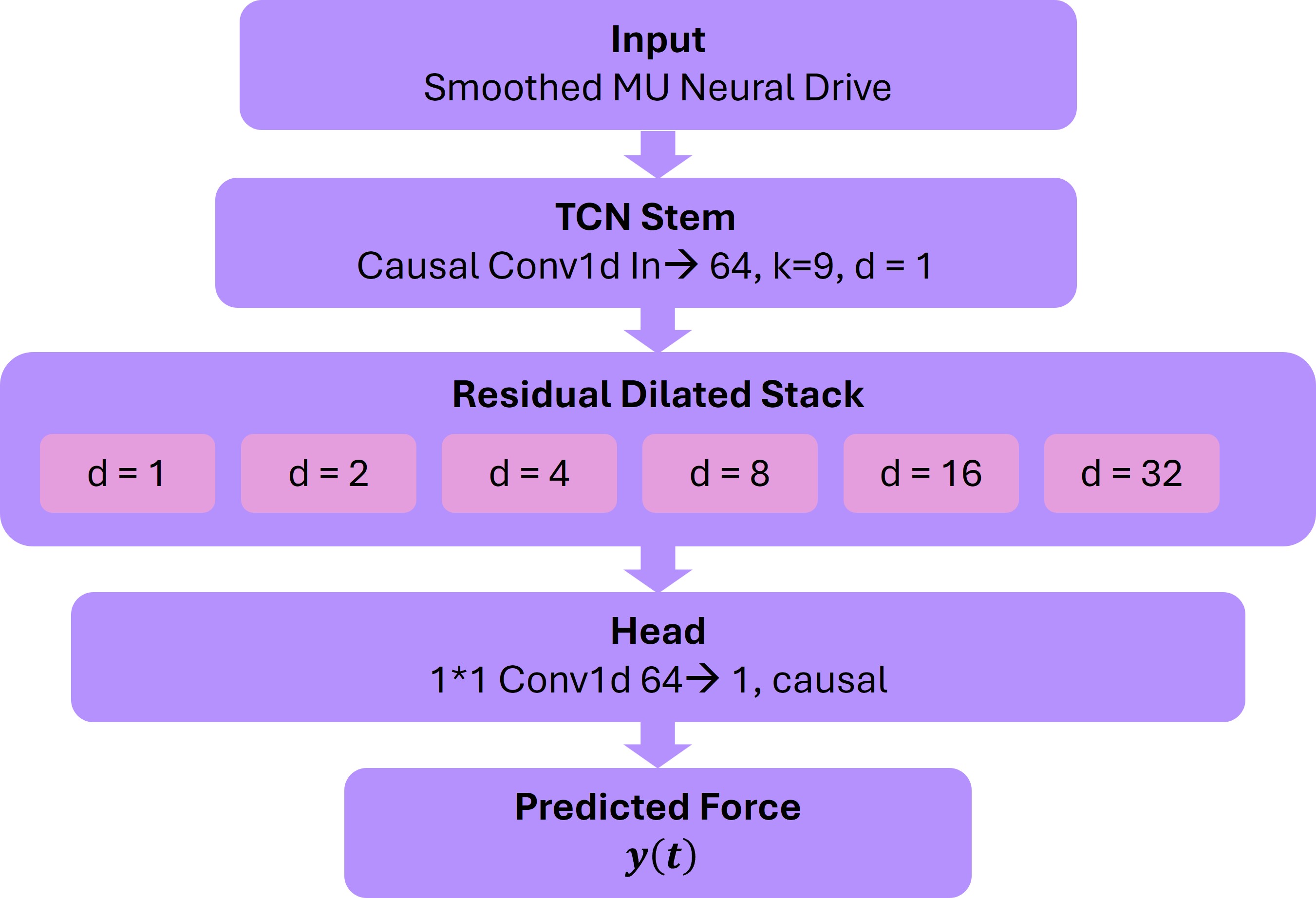}
  \caption{Temporal convolutional network block diagram (causal stem; six residual dilated blocks; \(1{\times}1\) head).}
  \label{fig:tcn}
\end{figure}

\begin{figure}[t]
  \centering
  \includegraphics[width=0.98\linewidth]{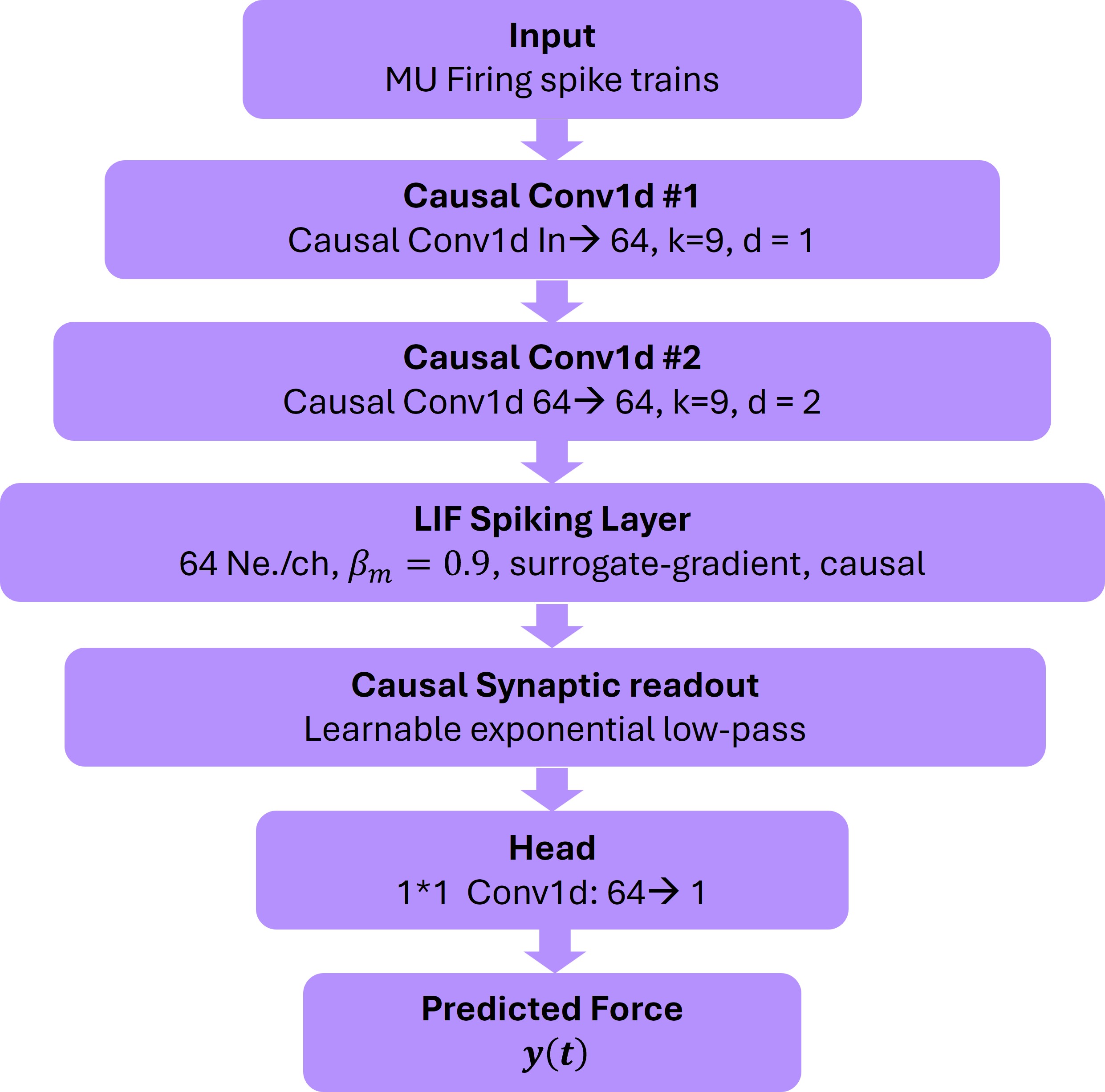}
  \caption{Spiking neural network block diagram (causal conv front-end; LIF layer; synaptic low-pass readout; \(1{\times}1\) head).}
  \label{fig:snn}
\end{figure}

\section{Results}
\begin{figure}[h]
  \centering
  \includegraphics[width=0.95\linewidth]{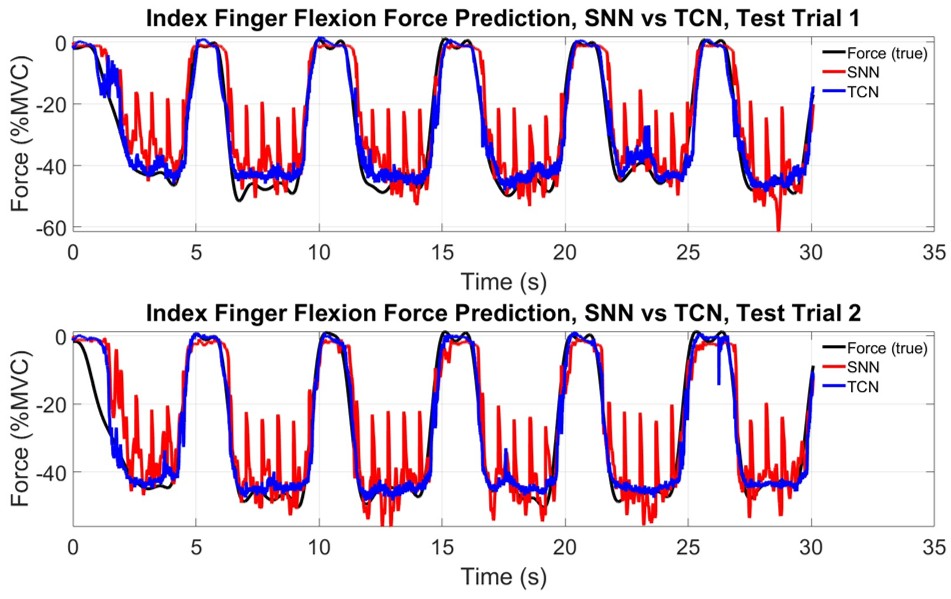}
  \caption{Predicted force of the models and measured force for two test trials.}
  \label{fig:traces}
\end{figure}

Across held-out trials, the TCN achieved lower error and higher correlation than the SNN baseline (Table~\ref{tab:metrics}). The predicted vs.\ measured force traces for both test trials are shown in Fig.~\ref{fig:traces}. Relative to the SNN, the TCN reduced RMSE by approximately \(46\%\) (from \(8.25\%\) MVF to \(4.44\%\) MVF), while both models achieved high correlations (\(r>0.92\)), indicating that each captured the overall temporal pattern of the force trajectory.

Visual inspection of Fig.~\ref{fig:traces} suggests that the TCN more closely tracks changes in force, particularly at higher force levels, whereas the SNN predictions exhibit large fluctuations and overshoots. This is consistent with the TCN’s larger effective receptive field and ability to combine information from multiple dilated scales, in contrast to the single LIF layer used in the current SNN configuration and the spiking nature of the SNN.

Across the two test trials, the ranking between models was consistent: TCN outperformed SNN on both RMSE and correlation, supporting the conclusion that the observed improvements are not driven by a single atypical trial but reflect a systematic advantage of the TCN architecture under the present training conditions.

\begin{table}[h]
\centering
\caption{Held-out performance (mean over test trials).}
\label{tab:metrics}
\begin{tabular}{lcc}
\toprule
Model & RMSE (\%MVF) & $r$ \\
\midrule
TCN & 4.44 & 0.974 \\
SNN & 8.25 & 0.922 \\
\bottomrule
\end{tabular}
\end{table}

\section{Discussion and Limitations}
TCN outperformed SNN in our assessment. The TCN’s dilated causal
convolutions capture long temporal context with stable gradient flow,
a behavior consistent with prior work showing that temporal
convolutional networks can be highly effective for sequence modeling
and regression tasks \cite{Bai2018_TCNSeqModel}.
Our SNN baseline used a single LIF layer without deeper spiking stacks
or additional filtering layers, limiting temporal expressivity and
showing spiky behavior; recent studies indicate that deeper spiking
architectures with residual connections can mitigate such limitations
and improve performance \cite{Neftci2019_SurrogateGrad,Fang2021_DeepResSNN}.

Despite lower accuracy, the SNN reached \(r \approx 0.92\), indicating a promising neuromorphic decoder with potential for low-power, real-time, and interpretable operation. Neuromorphic hardware
platforms such as Loihi and related CMOS neuromorphic systems have 
demonstrated substantial energy savings and low-latency inference for
spike-based workloads \cite{Davies2018_Loihi,Roy2019_NeuromorphicReview}.
In this context, the LIF membrane potentials and synaptic traces provide internal states that can, in principle, be inspected or constrained to match physiological time constants.  This is consistent with the role of
spiking neuron models as a bridge between biological and artificial computation \cite{Tanzarella2023_TNSRE,Maass1997_ThirdGenSNN}.

Adopting modest depth (2--3 LIF layers), a wider conv front-end, and
tuning of decay constants (\(\beta_m\), synaptic \(\alpha\)) are likely to close the gap. Previous work has shown that incorporating multi-timescale synapses or adaptive spiking units can extend the effective temporal memory of SNNs \cite{Yin2020_MultiTimescaleSRNN,Bellec2018_LSNN}, while residual connections and carefully designed surrogate-gradient training enable much deeper spiking architectures with stable optimization  \cite{Fang2021_DeepResSNN,Neftci2019_SurrogateGrad}. Targeted regularization of firing rates and sparsity can further balance accuracy and energy efficiency on neuromorphic hardware \cite{Roy2019_NeuromorphicReview}.

\textbf{Limitations \& future work.} This assessment used one subject and one task. Future work will extend to multi-participant datasets and additional tasks, and explore optimized neuromorphic architectures. In particular, it will be important to build subject-independent or subject-adaptive decoders, and compare energy and latency on neuromorphic hardware versus conventional GPUs/CPUs. Finally, the present analysis focused on force; extending the framework to multi-DOF hand kinematics and to participants post-stroke is a key next step toward clinically relevant assistive control.

\section{Conclusion}
We compared a neuromorphic SNN against a strong TCN baseline for fingertip-force decoding from HD-sEMG-derived MU activity. On held-out trials, the TCN attained lower error, while the SNN provided a credible, low-power baseline with clear avenues for improvement. The study demonstrates that, under a common HD-sEMG acquisition and MU-decomposition pipeline, dilated causal TCNs currently offer higher accuracy for continuous force regression, whereas SNNs already achieve respectable performance with a much simpler architecture.

These results establish a useful reference point for future neuromorphic decoders of motor intent and highlight the importance of architectural choices and temporal receptive fields in EMG-based regression. In future work, we plan to scale the approach to larger and more diverse datasets, refine SNN architectures and training protocols.

\section*{Artifacts}
The poster version of this work presented at IEEE EMBS NER 2025 is provided as an ancillary file on arXiv (\texttt{IEEE\_NER2025\_NeuromorphicEMG\_poster.pdf}).

\section*{Acknowledgments}
This work was supported by the National Institute on Disability, Independent Living, and Rehabilitation Research (ASSIST RERC, 90REGE0017-01) and (HCC\_2106747) from NSF. 

\bibliographystyle{IEEEtran}
\bibliography{refs}

\end{document}